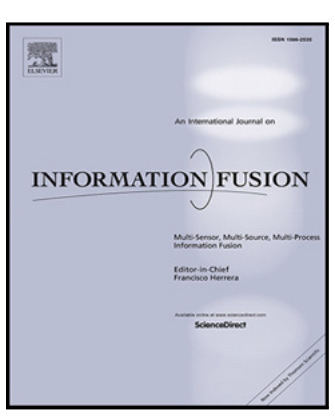

# Incomplete multimodal industrial anomaly detection via cross-modal distillation

Wenbo Sui [a,b,*], Daniel Lichau [a], Josselin Lefèvre [a,c], Harold Phelippeau [a]

[a] Thermo Fisher Scientific, Bordeaux, France
[b] Technical University of Denmark, Kgs. Lyngby, Denmark
[c] LIGM, Univ Gustave Eiffel, CNRS, Marne-la-Vallée, France

ARTICLE INFO

ABSTRACT

Recent studies of multimodal industrial anomaly detection (IAD) based on 3D point clouds and RGB images have highlighted the importance of exploiting the redundancy and complementarity among modalities. However, achieving multimodal IAD in practical production lines remains a work in progress as it considers the trade-offs between the costs and benefits associated with the introduction of new modalities, while ensuring compatibility with current processes. Existing quality control processes combine rapid in-line inspections, such as optical and ultrasound imaging with high-resolution but time-consuming near-line characterization techniques, like industrial CT and electron microscopy to manually or semi-automatically locate and analyze defects in the production of Li-ion batteries and composites. Given the cost and time limitations, only a subset of the samples can be inspected by all methods, and the remaining samples are only evaluated through one or two forms of in-line inspection. To fully exploit data for deep learning-driven defect detection, the models must have the ability to leverage multimodal training and handle incomplete modalities during inference. In this paper, we propose **CMDIAD**, a Cross-Modal Distillation framework for IAD to demonstrate the feasibility of a Multi-modal Training, Few-modal Inference (MTFI) pipeline. Our findings show that the MTFI pipeline can more effectively utilize incomplete multimodal information compared to applying only a single modality for training and inference. Moreover, we investigate the reasons behind the asymmetric improvement using point clouds or RGB images for inference. This provides a foundation for our future multimodal dataset construction with additional modalities from manufacturing scenarios. The code is released on https://github.com/evenrose/CMDIAD.

## 1. Introduction

Industrial anomaly detection (IAD) aims to detect abnormal characteristics of products during or after the manufacturing process thereby minimizing the risk of potential defects. The rapid development of deep learning-based image IAD has led to more accurate and robust detection and moved away from handcrafted feature engineering to learn more complex patterns for variable issues [1,2]. As a data-driven method, whether the model's input can provide sufficient defect-related information determines the upper limit of its performance [3,4]. As an example, 3D anomaly detection based on Point Clouds (PCs) input can capture unique spatial information of defects compared with RGB images but also loses sensitivity to color [5]. Therefore, taking advantage of the redundancy and complementary between modalities is crucial for IAD. The recent studies on multimodal IAD [6–8] demonstrate the superiority of multimodal input based on a benchmark dataset, MVTec 3D-AD [9], which introduced aligned 3D PCs and RGB images as complementary modalities.

Nevertheless, achieving multimodal IAD in practical production lines remains a work in progress. It requires consideration of the trade-offs between the costs and benefits associated with the introduction of new modalities, while ensuring compatibility with existing processes. For instance, the production of Li-ion batteries or composites integrates rapid in-line inspections, such as optical, infrared, X-ray and ultrasound imaging with high-resolution but time-consuming near-line characterization techniques, including industrial CT, X-ray scattering and electron microscopy to manually or semi-automatically analyze anomalies [10,11]. Due to cost and time constraints, only a subset of samples can be inspected by all in-line and near-line methods, and the remaining samples are evaluated by only one or two forms of in-line inspection. To fully exploit data for deep learning-driven automatic defect detection, the models must have the ability to leverage multimodal training and handle incomplete modalities during inference.

The research on anomaly detection in the presence of missing modalities is very limited [12]. A straightforward approach is to process the available modalities only with late fusion, such as the multi-memory bank method for unsupervised IAD [13] which can process the features of each modality independently for decision-making. However, when predicting on samples with incomplete modalities, it is not feasible to utilize the multimodal training set, as merely the network that processes the existing modality is operational. In the context of multivariate time series data with correlation between adjacent points, imputation can be employed to generate the value of the missing region [14–16], but the necessity for correlation also limits its applicability to non-time series data. Zhang et al. [17] utilizes a Generative Adversarial Network to impute the missing modality data using existing multimodal input. The complete multimodal data was then fed into the following models for anomaly detection, but the dual model structure will significantly increase the computational load.

* Corresponding author at: Technical University of Denmark, Kgs. Lyngby, Denmark.
*E-mail address:* websui@dtu.dk (W. Sui).

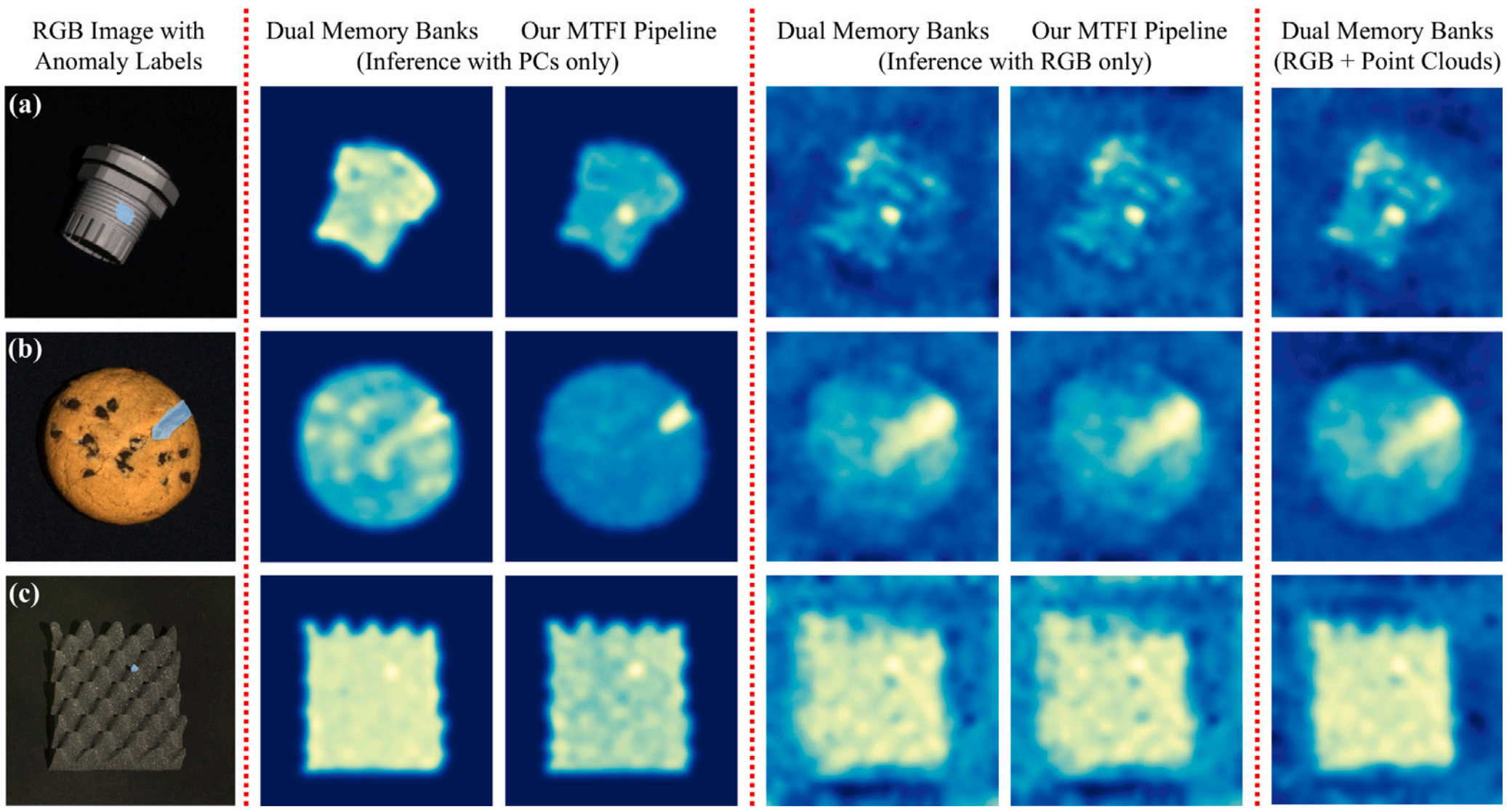

**Fig. 1.** Qualitative comparison of prediction results of Multi-modal Training, Few-modal Inference (MTFI) pipeline (Feature-to-Feature), dual memory banks (RGB+PCs for training and inference with RGB, PCs and RGB+PCs respectively) on: (a) Cable Gland-Thread-013, (b) Cookie-Crack-003 and (c) Foam-Contamination-01 of MVTec 3D-AD.

Researchers have tried to apply cross-modal Knowledge Distillation (KD) to deal with missing information more robustly in audio tagging [18], action recognition [19], medical image segmentation [20, 21], and saliency detection [22], etc. In contrast with the typical KD approaches, which involve teacher and student networks that accept the same modal input, the cross-modal KD's student network attempts to generate the same output as the teacher based on additional modalities. In other words, cross-modal KD is capable of estimating the missing information based on the known modalities by establishing a mapping between one modality and another. This inspires us to enhance the unsupervised IAD method through cross-modal KD to develop a Multi-modal Training, Few-modal Inference (MTFI) pipeline which enables the model to be trained with data from multiple inspection methods but perform inference with only one of them and benefit from the knowledge in the multimodal training set at the lowest cost.

In this paper, we propose **CMDIAD**, a Cross-Modal Distillation framework for IAD that learns to generate cross-modal hallucinations of missing modalities, with the objective of investigating the feasibility of the MTFI pipeline and what information can be transferred across modalities. We adopted the unsupervised multi-memory bank method from [8,13] as the foundation, with the capacity to process the features of each modality independently for the final decision. The core idea of our framework is to train learnable cross-modal KD networks to emulate the privileged information [23] like features or inputs of a given modality that is only present in the training and then try to generate its missing information during inference. In this study, three distinct distillation routes are proposed and evaluated in order to compare the increase in computational effort and the improvement in performance. To show their potential, we tested the framework feeding in 3D PCs information to generate RGB information on MVTec 3D-AD [9] and Eyecandies [24] datasets. The MTFI pipeline with Feature-to-Feature (F2F) network shows most promising improvement compared with the multi-memory bank method (dual memory banks specifically here, Fig. 1), and even exceeds the results of the SOTA Shape-guided model [25] with PCs input only, which combines PointNet [26] and Neural Implicit Function [27]. This once again shows the importance of multi-modal information and proves the feasibility of our pipeline. Interestingly, when we reverse the distillation direction and retrain the model to perform inference using RGB information and the PCs hallucination generated by the cross-modal KD networks, the performance improvement is either minimal or even declines. We further visualized and manually inspected the inputs and predictions, and attribute this asymmetry to the similar texture information between PCs and RGB images and the unique spatial information of PCs that cannot be estimated. These findings suggest that when constructing a multimodal IAD dataset in the future, it would be prudent to prioritize the use of in-line and near-line methods with associated information.

Following this introduction, Section 2 categorizes existing approaches of unsupervised 2D, 3D and multimodal IAD. Section 3 responds to the challenges by proposing a cross-modal distillation framework that enables multi-modal training, multi-modal inference workflows, and explains the networks and algorithms used. Experiment details and result discussion is conducted in Section 4. The paper concludes with limitations in Sections 5 and 6. The key contributions of our work are as follows:

- To the best of our knowledge, we are the first to implement IAD using incomplete multimodal data, which fits the actual scenario in the quality control process.
- We propose CMDIAD, a novel multimodal IAD framework which enables the model to be trained with multimodal data but perform inference with only one method.
- Our results illustrate the importance of associated information across modalities for incomplete multimodal IAD, and provide guidance for the selection of in-line and near-line methods for future dataset construction.

## 2. Related work

### 2.1. Unsupervised 2D industrial anomaly detection

Similar to other 2D anomaly detection fields that focus on unsupervised methods, such as medical [28] or hyperspectral AD [29], deep learning-driven IAD [1] faces the problem of insufficient prior knowledge of anomalies. To address this issue, different downstream tasks also adopted similar idea to maximize the distance of abnormal and normal input through feature embedding or reconstruction. The key of feature-embedding based methods is to translate normal samples into feature embeddings, and make a specific comparison among the generated features, *e.g.*, the teacher–student architecture [7,30] determines the anomaly scores by measuring the distance between the teacher and student network outputs which relies on the teacher pre-trained on large dataset. There are also similar methods through one-class classification [31,32] and feature distribution map [33,34]. Memory bank methods select and save the anomaly-free feature from the pre-trained network to form a memory, and then compare the features of test samples with the memory, which improves the recall [13,35,36]. Reconstruction based methods apply Auto-encoder (AE) [37,38], Generative Adversarial Networks (GAN) [39,40], or Diffusion Model [41, 42] and force the model to reconstruct normal samples. Since the model does not have the knowledge to reconstruct the abnormal part, the output of the abnormal sample will be partially dissimilar compared with the input, as a sign of anomalies. It is worth noting that although 2D IAD is limited by the amount of anomaly prior knowledge and data, there are few studies combining deep learning with explicit prior constraints compared to other tasks [43], which may be a way to solve overfitting and enhance generalization ability.

### 2.2. 3D and multi-modal RGB-3D industrial anomaly detection

The first public RGB-3D industrial anomaly detection dataset from real-world objects, MVTec 3D-AD [9], has greatly promoted studies in this field. AST [7] applied asymmetric teacher–student networks with RGB and depth map as inputs to avoid student learning anomalies through generalization. Horwitz and Hoshen [44] improved the utilization of spatial information by handcrafted and learned shape representations. A memory bank method, M3DM [8], replaced RGB and hand-crafted 3D features with frozen feature extractors pre-trained on large datasets like ImageNet [45] and generate fused features by forcing the output of additional learnable modules to be similar. Inspired by the adoption of feature extractors, Costanzino et al. [6] built a cross-modal mapping module and evaluated the anomaly score through comparing fake features with the real features. Since the features obtained from pre-trained extractors are independent which leads to difficulties in modality fusion, DADA [46] learned a joint representation of RGB-3D inputs by retraining the RGB backbone, resulting in enhanced detection performance. However, the above-mentioned RGB-3D IAD architectures require that the multi-modality of data must be complete during training and inference, and new architectures need to be developed to meet the needs of incomplete datasets in practical applications.

### 2.3. Cross-modal knowledge distillation

Most existing multimodal architectures require complete modalities for both training and inference, which limits their applicability in the real world with missing modalities [47]. Cross-modal KD is regarded as an effective method for missing modalities as its ability to transfer knowledge between different modalities. This method is also regarded as learning with privileged information in previous research [48,49], which means that some additional information (extra modalities) related to the training examples will be provided to the teacher, while this privileged information will not be available during inference. Garcia et al. [19] employed hallucination network as the student model to generate RGB-D cross-modal hallucination for video action recognition. Wang et al. [20] proposed LCKD to handle missing modalities through identifying the most important modality as a teacher. Previous research focused more on how to complete training in missing modalities with enhanced performance and lacked investigation into the correlation between modalities. However, inferring the latter through experimental results is more important because it can guide how to use existing modalities in quality control and the prior knowledge of human experts to improve the final defect detection rate of multimodal IAD.

## 3. Proposed method

Our Cross-Modal Distillation framework for IAD (CMDIAD) introduces a Multi-modal Training and Few-modal Inference (MTFI) pipeline as the basis of being compatible with single-modal training, single-modal inference and multi-modal training, multi-modal inference workflows. As shown in Fig. 2, this is achieved by: (1) frozen pre-trained networks for feature extraction of pre-processed RGB images and PCs; (2) coreset selection module for memory bank construction; (3) cross-modal distillation network to generate hallucination of one modality from another modality to compensate missing information during inference; (4) decision layer fusion module to utilize multiple memory banks for classification and segmentation.

### 3.1. Feature extraction

Two pre-trained transformers, DINO [50] and Point-MAE [51] are first applied as deep feature extractors ($\mathcal{F}_{RGB}$, $\mathcal{F}_{PC}$) following M3DM [8,52] for the contraction of memory bank due to their ability of producing distinct features between normal and anomalous data and handling variability of data. The input data of feature extractors and generated real features are denoted as ($I_{RGB}$, $I_{PC}$) and ($R_{RGB}$, $R_{PC}$), respectively.

#### 3.1.1. RGB feature extraction

For a given RGB image, $I_{RGB} \in \mathbb{R}^{H\times W\times 3}$, feature extractor $\mathcal{F}_{RGB}$ take it as input and produce a feature map $R_{RGB} \in \mathbb{R}^{H_f\times W_f\times d_1}$. To align feature maps from different modalities, we use up-pooling to obtain new $R_{RGB} \in \mathbb{R}^{2H_f\times 2W_f\times d_1}$. The positional encoding of the feature map is removed.

#### 3.1.2. Point clouds feature extraction and interpolation

For a given structured point cloud (PC) $I_{PC} \in \mathbb{R}^{H\times W\times 3}$, 3D local shape descriptors [51] are created as follows considering the inherent sparsity and disorder properties of point clouds. Farthest point sampling (FPS) [26] chooses $N$ points as group center for KNN clustering of $M$ points. Feature extractor $\mathcal{F}_{PC}$ further process $N$ point groups into feature embeddings with dimensions of $N\times d_2$. Then inverse distance weighting (IDW) method [8] is used for the interpolation of obtained features back to each pixel. The down sampled feature map $R_{PC} \in \mathbb{R}^{2H_f\times 2W_f\times d_2}$ is obtained through average pooling.

### 3.2. Cross-modal distillation

In order to implement the MTFI pipeline of fewer modalities inference, we build a cross-modal distillation network to generate hallucinations that retain the information of missing modalities. There are four possible distillation routes to satisfy the model: *(i)* Feature-to-Feature Distillation; *(ii)* Feature-to-Input Distillation; *(iii)* Input-to-Feature Distillation; and *(iv)* Input-to-Input distillation. Since *(iv)* still needs to extract features for estimation, which is essentially the same as *(ii)*, we only implemented *(i-iii)* for MTFI as shown in Fig. 2. The cross-modal distillation networks are denoted as $C_{F2F}$, $C_{Ft2I}$ and $C_{I2F}$.

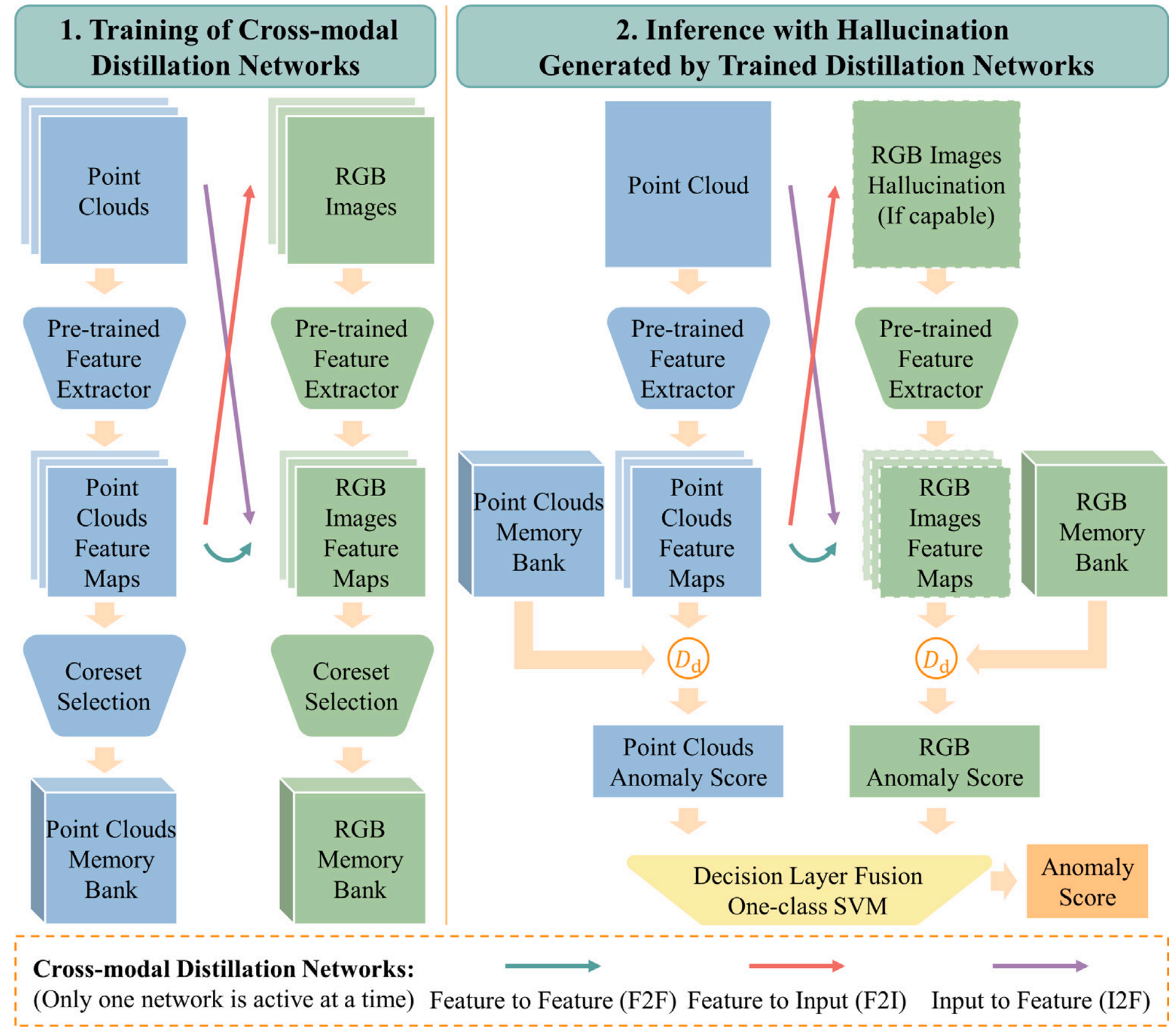

**Fig. 2.** A conceptual illustration of two stages of MTFI pipeline (Inference with PCs only).

*3.2.1. Feature-to-Feature (F2F) distillation*

Directly establishing a mapping from feature space to feature space is common and effective [6,19,20]. Even if the differences between modalities are large, the neural network can still learn common features or estimate new features. For a given feature $R^{i,j}_{PC}$ in a feature map $R_{PC}$, the estimated hallucination feature $H^{i,j}_{fRGB}$ can be obtained as follows:

$$H^{i,j}_{fRGB} = C_{F2F}(R^{i,j}_{PC}) \tag{1}$$

By generation of all features, we obtain the predicted feature maps $H^{i,j}_{fRGB}$ of dimensions $2H_f \times 2W_f \times d_1$. The estimation of $H^{i,j}_{PC}$ is similar based on the feature map $R^{i,j}_{fRGB}$.

*3.2.2. Feature-to-Input (F2I) distillation*

What type of knowledge can be retained during cross-modal distillation is unknown, so it is valuable to compare results with the feature extractor of missing modality or not. Estimating the input of one modality from the features (inputs) of another modality has been extensively studied as Single Image Depth Estimation (SIDS) or 3D Point Cloud Generation. The hallucination input $H_{iRGB} \in \mathbb{R}^{H\times W\times 3}$ and feature map $H_{fRGB} \in \mathbb{R}^{2H_f\times 2W_f\times d_1}$ are estimated based on the correspond feature map $R_{PC} \in \mathbb{R}^{2H_f\times 2W_f\times d_2}$ as:

$$H_{fRGB} = \mathcal{F}_{RGB}(H_{iRGB}), H_{iRGB} = C_{F2I}(R_{PC}) \tag{2}$$

*3.2.3. Input to Feature (I2F) distillation*

Similarly, it is also feasible to generate features of missing modalities through existing inputs as Eq. (3), which reduces the limitations of the feature extractor on the hidden layer and is closer to the teacher–student architecture of typical knowledge distillation. All background points in PCs input $I_{PC} \in \mathbb{R}^{H\times W\times 3}$ have been removed. For reverse distillation of $H_{fPC}$, the corresponding RGB points are also set to 0.

$$H_{fRGB} = C_{I2F}(I_{PC}) \tag{3}$$

*3.2.4. Optimization*

During the training, $C_{F2F}$, $C_{F2I}$ or $C_{I2F}$ are optimized based on all samples from anomaly-free training set by minimizing the average loss of per-pixel's distance between the real and hallucination, as Eqs. (4) and (5) (distillation RGB as an example). Considering the performance and computational cost, L2 distance is selected as the metric from L1, L2 and cosine distance Appendix A.1.

$$\mathcal{L}_{F2F\&I2F} = \sum_{i=1}^{H}\sum_{j=1}^{W} \|H^{i,j}_{fRGB} - R^{i,j}_{RGB}\|_2 \tag{4}$$

$$\mathcal{L}_{F2I} = \sum_{i=1}^{H}\sum_{j=1}^{W} \|H^{i,j}_{iRGB} - I^{i,j}_{RGB}\|_2 \tag{5}$$

### 3.3. Coreset selection for memory bank construction

The memory bank method determines the anomaly score by measuring the distance between the sample features to be tested and the normal features in the memory bank. Thus, it is crucial to make the memory bank as comprehensive as possible and remove duplicate features to reduce the calculation cost instead of storing all features as $\mathcal{M}$ for inference. To maximize the representative of the memory bank $\mathcal{M}_{coreset}$, the solution of an NP-Hard problem [53] needs to be found:

$$\mathcal{M}_{coreset} = \arg \min_{\mathcal{M}_{coreset}\in\mathcal{M}} \max_{m\in\mathcal{M}} \min_{n\in\mathcal{M}_{coreset}} D_c(m,n) \tag{6}$$

A greedy algorithm for sampling is used similar to PatchCore [13] to address this problem. Before the selection of core patches, all patches in the memory bank are first normalized with modality-specific mean and standard deviation to suppress the effect of variability. Starting from a random given patch embedding $p_i \in \mathbb{R}^{1\times d}$ from a feature set $S \in \mathbb{R}^{P\times d}$ with $P$ patch features, a new selected patch $p_{i+1}$ is calculated as follows:

$$p_{i+1} = \arg \max_{p_j\in S, i\neq j} D_c(p_i, p_j) \tag{7}$$

$D_c$ is the distance metric which needs to be consistent with the one used to calculate the anomaly score in Section 3.4. The memory banks $\mathcal{M}_{PC}, \mathcal{M}_{RGB}$ in Fig. 2 are constructed by adding the selected patch to memory and calculate the new patch through $p_{i+1}$ iteratively for limited times. However, as the number of samples increases, the process of calculating high dimensional distance is still time-consuming. Sparse Random Projection [8,54] is applied to reduce the dimension from $d$ to $d'$ ($d' \ll d$) and accelerate computation with acceptable embedding quality. The optimized time complexity of the whole coreset selection is dominated by $O(Pdd')$ for projection plus $O(P^2 d')$ for distance calculation ($P$ is the sample number of coreset).

### 3.4. Decision layer fusion

To make the model more lightweight, late fusion is deployed. At inference time, the model calculates anomaly scores by comparing each real or generated feature map $F_{PC}$ and $F_{RGB}$ with correspond memory bank. Two One-Class Support Vector Machine using Stochastic Gradient Descent, $C_c$ and $C_s$, are applied as the decision layer fusion for image classification and segmentation as:

$$c = C_c(\alpha\psi(F_{PC}, \mathcal{M}_{PC}),\ \beta\psi(F_{RGB}, \mathcal{M}_{RGB})) \tag{8}$$

$$s = C_s(\alpha\phi(F_{PC}, \mathcal{M}_{PC}),\ \beta\phi(F_{RGB}, \mathcal{M}_{RGB})) \tag{9}$$

c and s are final predication score of image classification and pixel segmentation. In order to balance the mean of anomaly scores between modalities, correction factor $\alpha$ and $\beta$ are introduced. $\psi$ and $\phi$ are anomaly scores of classification and segmentation from [13] formulated as:

$$\psi(F, \mathcal{M}) = D_d(F^{(i,j),*}, m^*) \tag{10}$$

$$\phi(F, \mathcal{M}) = \{\min_{m\in\mathcal{M}} D_d(F^{i,j}, m),\ for\ F^{i,j} \in F\} \tag{11}$$

$D_d$ is the distance metric. The anomaly score of a image is determined by the maximum distance of between a feature and its most similar feature $m$ in memory bank:

$$F^{(i,j),*}, m^* = \arg\max_{F^{(i,j)}\in F} \arg\min_{m\in\mathcal{M}} D_d(F^{i,j}, m) \tag{12}$$

## 4. Experiments

### 4.1. Experimental setup

#### 4.1.1. Dataset and evaluation metrics

Majority of recent studies in multimodal IAD research are on the MVTec 3D-AD [9] dataset, which includes aligned PCs and RGB images of real-world objects. This dataset consists of 10 classes of objects and each class may have 3–5 types of defects with pixel-level binary ground truth. Objects were recorded with a structured light 3D sensor. Every sample contains one image with (x, y, z) values of corresponding PCs and another image with (r, g, b) values. We also evaluate our methods on Eyecandies [24], a similar synthetic aligned RGB+3D IAD dataset. The image and pixel level anomaly detection performance are evaluated with Area Under the Receiver Operator Curve (I-AUROC and P-AUROC) and Area Under the Per-Region Overlap (AUPRO) following [8,9]. I-AUROC focuses on whether the entire image contains anomalies, while P-AUROC focuses on the model's ability to identify anomalies at the pixel level. AUPRO is the average overlap of the prediction with each connected component of ground truth, thereby reducing the impact of the size of anomalies compared with P-AUROC.

#### 4.1.2. Data pre-processing

To solve the classification errors caused by background artifacts, the background plane of PCs was estimated based on the RANSAC [55] method following [8,44]. Then we replaced the values of points within 0.005 from the plane with zeros.

**Table 1**
Parameter numbers and FLOPs of used feature extractors and cross-modal distillation networks.

| Networks | #Params (M) | FLOPs (G) |
|---|---|---|
| RGB feature extractor | 85.204 | 66.850 |
| PCs feature extractor | 21.825 | 86.645 |
| F2F | 6.642 | 20.819 |
| F2I | 3.015 | 26.401 |
| I2F | 10.215 | 32.175 |

#### 4.1.3. Implementation details

Two feature extractors DINO [50] and Point-MAE [51] pre-trained on ImageNet [45] and ShapeNet [56] are used as [8]. The parameters of the feature extractor and the floating point operations (FLOPs) are shown in Table 1. For an input RGB image, (224, 224, 3), the corresponding feature map is up-sampled to (56 × 56, 768). For the PCs, FPS is first used to convert it into 1024 groups and each group has 128 points (1024, 384) as the input of point transformer. The features generated by the point transformer are interpolated into a (56 × 56, 768) feature map through the inverse distance weight method, yielding aligned RGB and PCs feature maps. The learning rate of One-Class Support Vector Machine using Stochastic Gradient Descent for decision layer fusion is set as $1 \times 10^{-4}$ for 1000 steps.

For three levels of cross-modal distillation, three different networks are employed to provide suitable fitting capabilities in accordance with previous studies. For F2F distillation, a three-layer MLP with (1920, 1920, 768) units is used to generate the hallucination of cross-modal feature. For I2F distillation, we adopted the high-resolution branch of HRNet [57] to retain more information and generate high-resolution feature maps. A convolutional network similar to the depth estimation head of DINOv2 [58] was used for F2I distillation. Similarly, the parameter numbers and FLOPs of the distillation networks are shown in Table 1. Adam optimizer was applied with a batch size of 32 and 10 epochs warm-up to train all three types of networks for 100 epochs, and early stop based on the results on the evaluation set. We report the best results of F2F and F2I distillation setting the learning rate as 0.0005 at 64/39 (PCs) and 49/54 (RGB) epochs respectively. For I2F distillation, we reduce the learning rate to 0.0003 to avoid divergence and obtain the optimal results from 24 epochs for PCs and 34 epochs for RGB images. The best epochs are selected by performing inference on the validation set with trained distillation network.

### 4.2. Results of MTFI pipeline (inference with PCs)

Table 2 shows the I-AUROC and AUPRO of our method and previous work for detecting anomalies on the 10 classes of MVTec 3D-AD and the averages. The P-AUROC is also attached in Appendix A.1. We compared F2F, F2I and I2F of our MTFI pipeline with dual memory banks inference with PCs only (DualBanksPCs) and other methods based on pure PCs or depth images (For details about others methods, please refer to Section 2). Since only one memory bank is activated during the inference of DualBanksPCs, the result is equivalent to that of the method using a single memory bank, and a noticeable performance degradation (I-AUROC 7.8% ↓, AUPRO 5.1% ↓) can be observed compared with the complete dual memory bank method (DualBanksRGB+PCs, Table 4). For MTFI pipeline, both RGB and PCs data were utilized during training, while solely PCs data was employed during the inference. Both F2F and F2I are substantially better than memory bank method (F2F: I-AUROC 7.8% ↑, AUPRO 2.3% ↑; F2I: I-AUROC 0.3% ↑, AUPRO 0.1% ↑). The highest I-AUROC result is from our MTFI pipeline (F2F, I-AUROC 0.938, AUPRO 0.934), which outperforms the SOTA work, Shape-guided [25] with specially designed memory bank for PCs by 2.2% and 0.3% for I-AUROC and AUPRO, respectively. This demonstrates the feasibility of cross-modal distillation and the MTFI pipeline for IAD. F2F distillation also has the same performance in image classification as the dual memory banks inference with RGB+PCs, and has faster inference speed (comparison in Table 5 and Appendix A.6). We attribute the performance gap between F2F and other distillation networks to insufficient data, since F2I and I2F involve feature extraction or image generation, which require a large amount of data to learn complex patterns, considering that we have allocated the fitting capacity to different paths according to experience and literature.

**Table 2**
I-AUROC and AUPRO results on MVTec 3D-AD for 3D based methods.

| | Method | Bagel | CableGland | Carrot | Cookie | Dowel | Foam | Peach | Potato | Rope | Tire | Mean |
|---|---|---|---|---|---|---|---|---|---|---|---|---|
| I-AUROC | 3D-ST [5] | 0.862 | 0.484 | 0.832 | 0.894 | 0.848 | 0.663 | 0.763 | 0.687 | 0.958 | 0.486 | 0.748 |
| | FPFH [44] | 0.825 | 0.551 | 0.952 | 0.797 | 0.883 | 0.582 | 0.758 | 0.889 | 0.929 | 0.653 | 0.782 |
| | AST [7] | 0.881 | 0.576 | 0.965 | 0.957 | 0.679 | 0.797 | 0.990 | 0.915 | 0.956 | 0.611 | 0.833 |
| | M3DM [8] | 0.941 | 0.651 | 0.965 | 0.969 | 0.905 | 0.760 | 0.880 | 0.974 | 0.926 | 0.765 | 0.874 |
| | Shape-guided [25] | 0.983 | 0.682 | **0.978** | **0.998** | 0.960 | 0.737 | **0.993** | **0.979** | **0.966** | 0.871 | 0.916 |
| | Ours$_{DualBanksPCs}$ | 0.973 | 0.687 | 0.927 | 0.965 | 0.838 | 0.732 | 0.857 | 0.987 | 0.864 | 0.863 | 0.860 |
| | Ours$_{FtoF}$ | **0.992** | **0.893** | 0.977 | 0.960 | 0.953 | **0.883** | 0.950 | 0.937 | 0.943 | **0.893** | **0.938** |
| | Ours$_{FtoI}$ | 0.976 | 0.780 | 0.974 | 0.810 | 0.879 | 0.735 | 0.929 | 0.877 | 0.855 | 0.815 | 0.863 |
| | Ours$_{ItoF}$ | 0.981 | 0.719 | 0.842 | 0.963 | **0.965** | 0.437 | 0.815 | 0.816 | 0.875 | 0.892 | 0.820 |
| AUPRO | Voxel VM [9] | 0.453 | 0.343 | 0.521 | 0.697 | 0.680 | 0.284 | 0.349 | 0.634 | 0.616 | 0.346 | 0.492 |
| | FPFH [44] | 0.973 | 0.879 | **0.982** | 0.906 | 0.892 | 0.735 | 0.977 | 0.982 | 0.956 | 0.961 | 0.924 |
| | M3DM [8] | 0.943 | 0.818 | 0.977 | 0.882 | 0.881 | 0.743 | 0.958 | 0.974 | 0.950 | 0.929 | 0.906 |
| | Shape-guided [25] | **0.974** | 0.871 | 0.981 | 0.924 | 0.898 | 0.773 | 0.978 | **0.983** | 0.955 | 0.969 | 0.931 |
| | Ours$_{DualBanksPCs}$ | 0.947 | 0.826 | 0.977 | 0.882 | 0.881 | 0.767 | 0.967 | 0.978 | 0.947 | 0.940 | 0.911 |
| | Ours$_{FtoF}$ | 0.968 | 0.907 | **0.982** | **0.933** | 0.918 | 0.726 | **0.982** | **0.983** | **0.969** | **0.974** | 0.934 |
| | Ours$_{FtoI}$ | 0.963 | 0.826 | 0.978 | 0.879 | **0.947** | 0.709 | 0.972 | 0.965 | 0.951 | 0.940 | 0.912 |
| | Ours$_{ItoF}$ | 0.947 | **0.973** | 0.973 | 0.893 | 0.935 | **0.843** | 0.970 | 0.956 | 0.964 | 0.968 | **0.942** |

**Table 3**
I-AUROC and AUPRO results on MVTec 3D-AD for RGB based methods.

| | Method | Bagel | CableGland | Carrot | Cookie | Dowel | Foam | Peach | Potato | Rope | Tire | Mean |
|---|---|---|---|---|---|---|---|---|---|---|---|---|
| I-AUROC | PADiM [59] | **0.975** | 0.775 | 0.698 | 0.582 | 0.959 | 0.663 | 0.858 | 0.535 | 0.832 | 0.760 | 0.764 |
| | PatchCore [13] | 0.876 | 0.880 | 0.791 | 0.682 | 0.912 | 0.701 | 0.695 | 0.618 | 0.841 | 0.702 | 0.770 |
| | STFPM [60] | 0.930 | 0.847 | 0.890 | 0.575 | 0.947 | 0.766 | 0.710 | 0.598 | 0.965 | 0.701 | 0.793 |
| | CS-Flow [61] | 0.941 | 0.930 | 0.827 | 0.795 | **0.990** | 0.886 | 0.731 | 0.471 | 0.986 | 0.745 | 0.830 |
| | AST [7] | 0.947 | 0.928 | 0.851 | **0.825** | 0.981 | **0.951** | 0.895 | 0.613 | **0.992** | **0.821** | **0.880** |
| | M3DM [8] | 0.944 | 0.918 | 0.896 | 0.749 | 0.959 | 0.767 | 0.919 | **0.648** | 0.938 | 0.767 | 0.850 |
| | Shape-guided [25] | 0.911 | **0.936** | 0.883 | 0.662 | 0.974 | 0.772 | 0.785 | 0.641 | 0.884 | 0.706 | 0.815 |
| | Ours$_{DualBanksRGB}$ | 0.942 | 0.918 | 0.896 | 0.749 | 0.959 | 0.767 | 0.919 | **0.648** | 0.941 | 0.768 | 0.851 |
| | Ours$_{FtoF}$ | 0.945 | 0.926 | 0.896 | 0.700 | 0.965 | 0.836 | 0.925 | 0.644 | 0.942 | 0.777 | 0.856 |
| | Ours$_{FtoI}$ | 0.918 | 0.900 | 0.859 | 0.750 | 0.933 | 0.781 | 0.832 | 0.574 | 0.934 | 0.744 | 0.823 |
| | Ours$_{ItoF}$ | 0.938 | 0.929 | **0.897** | 0.760 | 0.959 | 0.802 | **0.932** | 0.638 | 0.941 | 0.787 | 0.858 |
| AUPRO | CS-Flow [61] | 0.855 | 0.919 | 0.958 | 0.867 | **0.969** | 0.500 | 0.889 | 0.935 | 0.904 | 0.919 | 0.871 |
| | PatchCore [13] | 0.901 | 0.949 | 0.928 | 0.877 | 0.892 | 0.563 | 0.904 | 0.932 | 0.908 | 0.906 | 0.876 |
| | PADiM [59] | **0.980** | 0.944 | 0.945 | 0.925 | 0.961 | 0.792 | 0.966 | 0.940 | 0.937 | 0.912 | 0.930 |
| | M3DM [8] | 0.952 | 0.972 | 0.973 | 0.891 | 0.932 | 0.843 | **0.970** | 0.956 | 0.968 | 0.966 | 0.942 |
| | Shape-guided [25] | 0.946 | **0.972** | 0.960 | **0.914** | 0.958 | 0.776 | 0.937 | 0.949 | 0.956 | 0.957 | 0.933 |
| | Ours$_{DualBanksRGB}$ | 0.951 | **0.972** | **0.973** | 0.891 | 0.932 | 0.843 | **0.970** | **0.956** | **0.968** | 0.966 | 0.942 |
| | Ours$_{FtoF}$ | 0.943 | 0.971 | **0.973** | 0.879 | 0.936 | 0.845 | 0.968 | **0.956** | **0.968** | 0.965 | 0.940 |
| | Ours$_{FtoI}$ | 0.897 | 0.946 | 0.923 | 0.842 | 0.753 | 0.805 | 0.927 | 0.880 | 0.917 | 0.930 | 0.882 |
| | Ours$_{ItoF}$ | 0.950 | **0.972** | **0.973** | 0.893 | 0.932 | **0.849** | **0.970** | **0.956** | **0.968** | **0.968** | **0.943** |

### 4.3. Results of MTFI Pipeline (Inference with RGB)

Exceedingly different results were obtained from MTFI Pipeline (Inference with RGB only) and shown in Table 3. Only slight improvement in F2F and I2F can be observed with dual memory bank method with RGB input only (DualBanksRGB). This asymmetry result indicates that the generated PCs hallucination cannot provide enough information to distinguish between normal and abnormal patches for decision layer fusion.

### 4.4. Ablation study

We conduct an ablation study to verify the effectiveness of MTFI pipeline and assess the generalizability of asymmetric performance by changing the feature extractor and dataset. We replace the used RGB feature extractor DINO ViT-B/8 with ViT-B/8-in21k and DINO ViT-S/8, PCs feature extractor Point-MAE with Point-Bert, and dataset MVTec 3D-AD with Eyecandies respectively (Table 4). The results based on the MTFI pipeline (inference with PCs) are markedly superior to those based on DualBanksPCs, and even better than dual RGB+PCs memory banks with complete input modalities in some cases. But the MTFI pipeline (inference with RGB) is barely an improvement over DualBanksRGB. There is a clear difference in the effects of trained and untrained cross-modal distillation networks Appendix A.2. All those results further demonstrate that MTFI pipeline (inference with PCs) can enhance the utilization of multimodal training data. Furthermore, the coreset selection ratio is also evaluated based on I-AUROC, AUPRO and the Inference time in Table 5. A ratio of 10% shows the best performance, while a larger ratio does not improve the performance and significantly reduces the inference speed.

### 4.5. Visualization and analysis of classification results of MTFI pipeline

To provide an intuitive understanding of the differences in detection results, we visualized and manually inspected the inputs and predictions. The Cable Gland class was investigated first for the dramatically enhanced performance which is mainly attributed to the reduction of false positives and correct identification of ‘thread’ anomalies like Fig. 1a. ‘Thread’ class has smaller local shape changes in PCs compared to the other classes ‘bent’, ‘cut’ and ‘hole’ where large chunks of structure are obviously bent or missing (Figs. 3, 4). In conventional RGB feature extraction, color and texture are generally accepted to be the most effective features for classification [44,62] and geometry texture information can also be extracted from PCs [63,64] through changes in local shape. For ‘thread’, local texture disappears due to damage, which is vaguely visible in the PCs and is confused by the model as a normal feature. But the texture of this region is quite different from the of normal area in the RGB image. For cross-modal KD, the RGB hallucination generated through PCs preserve this texture abnormality and provide discrimination. This is evidenced by the improved classification of ‘crack’ in Cookie class and ‘contamination’ in Foam class. There are many pits on the cookie surface that the PCs model considers to be similar to cracks (Fig. 1b) and the undulating structure of the foam surface is indistinguishable from the contamination (Fig. 1c). This limited local shape difference cannot provide specific abnormal information, but the generated RGB hallucination is obviously different from the normal area. In short, the MTFI pipeline enables features that are visible in both modalities but insufficient for accurate discrimination in the original modality to be accurately classified, potentially through mapping shared or similar information among modalities like texture.

**Table 4**
Ablation study for different feature extractors on MVTec 3D-AD and replacing dataset with Eyecandles. The top and bottom of each row are the average I-AUROC and AUPRO. RGB#1: DINO ViT-B/8, RGB#2: DINO ViT-S/8, RGB#3: ViT-B/8-in21k, PCs#1: Point-MAE, PCs#2: Point-Bert. RGB#1 and PCs#1 are used in previous MTFI study.

| Feature extractors | DualBanks PCs | MTFI PCs | DualBanks RGB | MTFI RGB | DualBanks RGB+PCs |
|---|---|---|---|---|---|
| RGB#1+PCs#1 | 0.860<br>0.911 | 0.938<br>0.934 | 0.851<br>0.942 | 0.856<br>0.940 | 0.938<br>0.962 |
| RGB#2+PCs#1 | 0.860<br>0.911 | 0.899<br>0.931 | 0.817<br>0.937 | 0.817<br>0.938 | 0.908<br>0.911 |
| RGB#3+PCs#1 | 0.860<br>0.911 | 0.884<br>0.920 | 0.663<br>0.837 | 0.670<br>0.838 | 0.873<br>0.911 |
| RGB#1+PCs#2 | 0.790<br>0.869 | 0.864<br>0.926 | 0.851<br>0.942 | 0.860<br>0.943 | 0.833<br>0.948 |
| RGB#1+PCs#1(Eyecandies) | 0.732<br>0.684 | 0.764<br>0.726 | 0.869<br>0.884 | 0.874<br>0.884 | 0.859<br>0.881 |

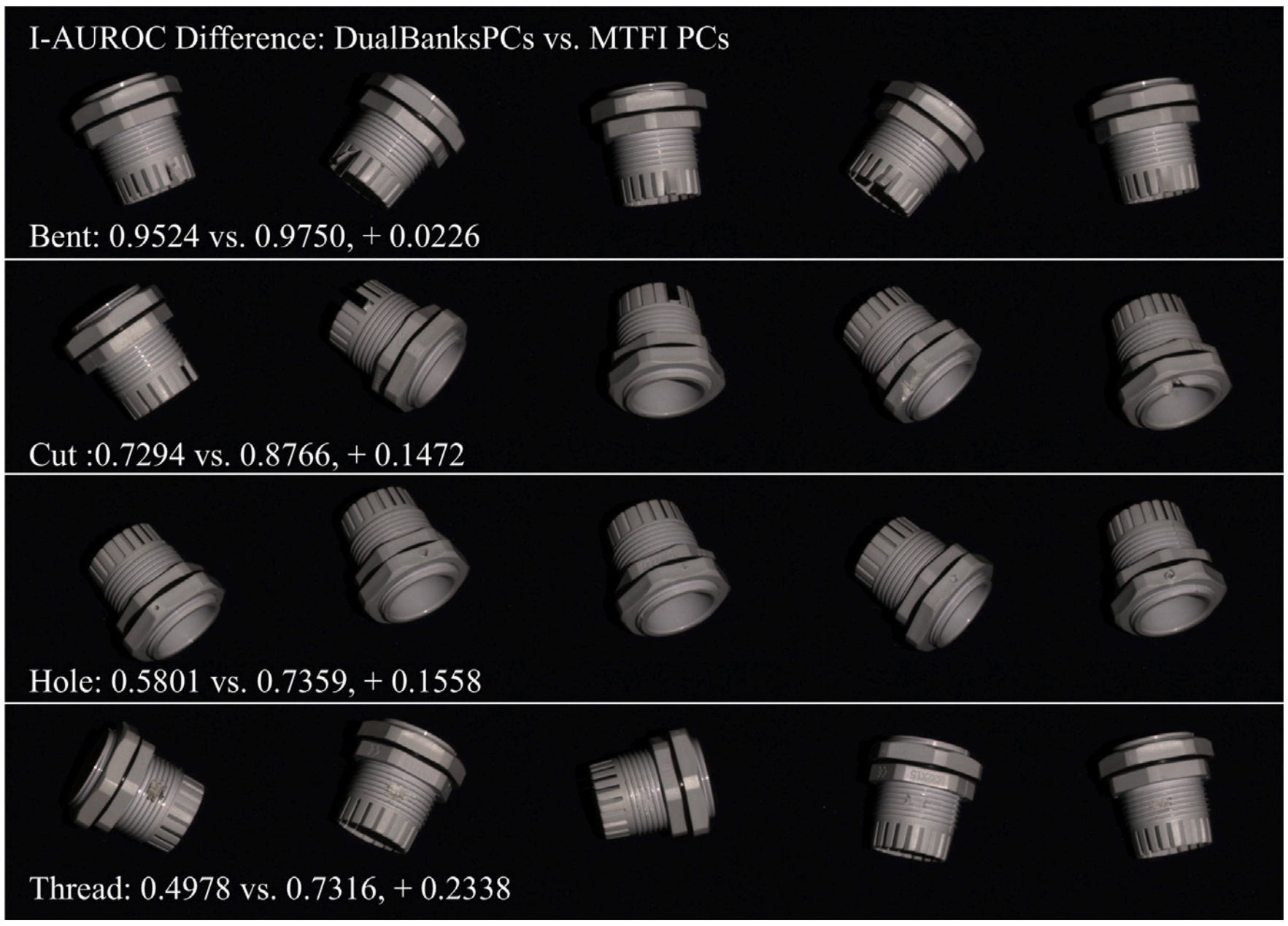

**Fig. 3.** Visualization of ‘bent’, ‘cut’, ‘hole’ and ‘thread’ anomalies of Cable Gland class.

**Table 5**
Comparison of different Coreset selection ratio.

| Coreset ratio (%) | 1 | 2 | 5 | 10 | 20 |
|---|---|---|---|---|---|
| I-AUROC | 0.929 | 0.928 | 0.930 | 0.938 | 0.935 |
| AUPRO | 0.930 | 0.932 | 0.934 | 0.934 | 0.934 |
| Inference time (s) | 1.02 | 1.07 | 1.35 | 1.74 | 2.90 |

### 4.6. *Analysis of asymmetry results of cross-modal distillation*

However, the results of MTFI pipeline (inference with RGB) reveal that anomalies visible in both modalities is only one sufficient condition. Comparing MTFI pipeline with PCs and RGB images from Cable Grand class as an example, anomalies that cannot be accurately classified using RGB images are mainly from the ‘bent’ anomaly, *e.g.*, Bent-8 and Bent-10 in Fig. 4e, f. This bending of part of the structure is very similar to other structures in RGB images. It needs to be classified through spatial information (shape features), which is what PCs are better at. Previous research has proven that it is challenging to extract shape or spatial features from a single RGB image, especially with limited data [65,66]. We speculate that the asymmetry results due to the insufficient spatial information in RGB modality, which is necessary for accurate PC hallucination generation to distinguish normal and abnormal inputs. This is also mirrored in almost the same ranking of sample anomaly score from MTFI pipeline (inference with RGB) and DualBanksRGB. Results of ‘hole’ and ‘combined’ from Cookie class and ‘cut’ from Foam class are consistent with the assumption of lacking spatial information Appendix A.3. In summary, two points should be noted to enhance single-modal inference through MTFI pipeline: (1) anomalies must be visible in both modalities but can be insignificant in one modality for classification. (2) shared information like texture or spatial information needs to be both present and represented in latent space to help generate hallucination with abnormality. Although complex representations are challenging for humans to comprehend and, the inability to generate valuable PCs hallucination using RGB that lacks spatial information highlights the significance of shared information.

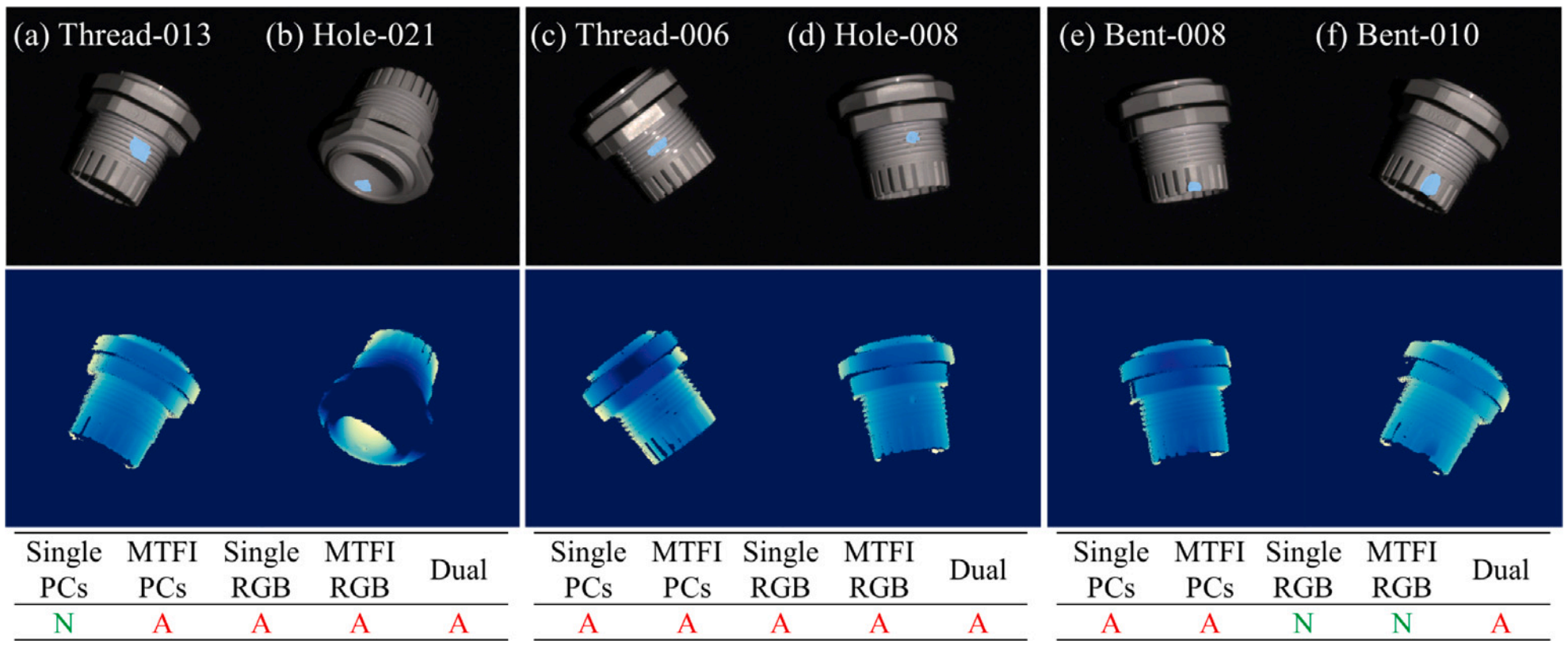

**Fig. 4.** Visualization of 6 representative abnormal samples with image classification results (A: Abnormal, N: Normal). (a, b) require cross-modal distillation to aid classification due to small shape changes. (c, d) can be directly classified by DualBanksPCs due to obvious shape changes. (e, f) can only be correctly classified in the PCs based method, although the structure changes obviously, the anomalies are too similar to other structures in the RGB image.

## 5. Conclusion

In this work, we have developed an effective framework based on MTFI pipeline driven via cross-modality distillation for multimodal IAD. It can retain partial information of the missing modality by generating its hallucination during inference. The MTFI pipeline (inference with PCs) delivers more accurate predictions compared with DualBanksPCs, but the MTFI pipeline (inference with RGB) only has a moderate positive effect. Thanks to the lightweight FtoF cross-modal distillation network, the inference speed of the MTFI pipeline is faster than that of the dual memory bank method and is similar to that of a single memory bank. This architecture also makes it possible to expand to more than two modalities by simply transforming the objective of the cross-modal distillation network from a one-to-one feature mapping to completing the full multimodal features. By manually inspecting the variation in classification results, we attribute this to the inability of generating PCs hallucination with abnormality from RGB for lacking of spatial information. For anomalies that are confused by the DualBanksPCs due to limited shape changes in the 3D structure, the RGB hallucination generated through modal shared information like texture provides effective discrimination. This indicates that the compatibility modalities needs to be considered in the our future multi-modal IAD dataset collection.

## 6. Limitations and future

In addition, we have demonstrated the promising results of CMDIAD based on the MTFI pipeline on two datasets and multiple feature extractors and its scalability, but the use of pre-trained feature extractors limits the extension of the method to some modalities that do not have large-scale datasets, such as ultrasound, infrared, etc. Further investigation will attempt to remove the dependency with transfer learning [29]. In the absence of feature extractors pre-trained on large datasets, one should also consider how to explicitly handle data variability as [67]. On the other hand, the application of the MTFI pipeline in real scenarios still requires the collection of a large amount of registered in-line and near-line data, which is also our next priority.

## CRediT authorship contribution statement

**Wenbo Sui:** Writing – review & editing, Writing – original draft, Visualization, Methodology, Conceptualization. **Daniel Lichau:** Writing – review & editing, Methodology. **Josselin Lefèvre:** Writing – review & editing, Methodology. **Harold Phelippeau:** Writing – review & editing, Project administration, Conceptualization.

## Declaration of competing interest

The authors declare that they have no known competing financial interests or personal relationships that could have appeared to influence the work reported in this paper.

## Acknowledgment

The authors acknowledge funding from Horizon Europe through the MSCA Doctoral Network RELIANCE, grant no. 101073040.

## Appendix

### A.1. Additional experiment results on MVTec 3D-AD

The experimental data not included in the main text are attached here, P-AUROC results on MVTec 3D-AD (Table A.1) and I-AUROC and AUPRO results on MVTec 3D-AD for RGB+PCs based methods (Table A.3). Although many studies still report P-AUROC, the comparative value is low because the values are very high and close.

### A.2. Additional experiment results of ablation study

As a supplement to the ablation experiment, a detailed comparison of teacher model and trained/untrained student model (F2F) is shown in Table A.2.

In order to compare the efficiency of utilizing multimodal data, we also tested the basic dual memory bank method shown in Table A.3.

The experimental results on the distance metrics (L1, L2, Cosine) are shown in Table A.4.

### A.3. Supplementary prediction results of samples in Fig. 4

The prediction results of the six samples in Fig. 4 are shown in Figs. A.1–A.6 as a supplement.

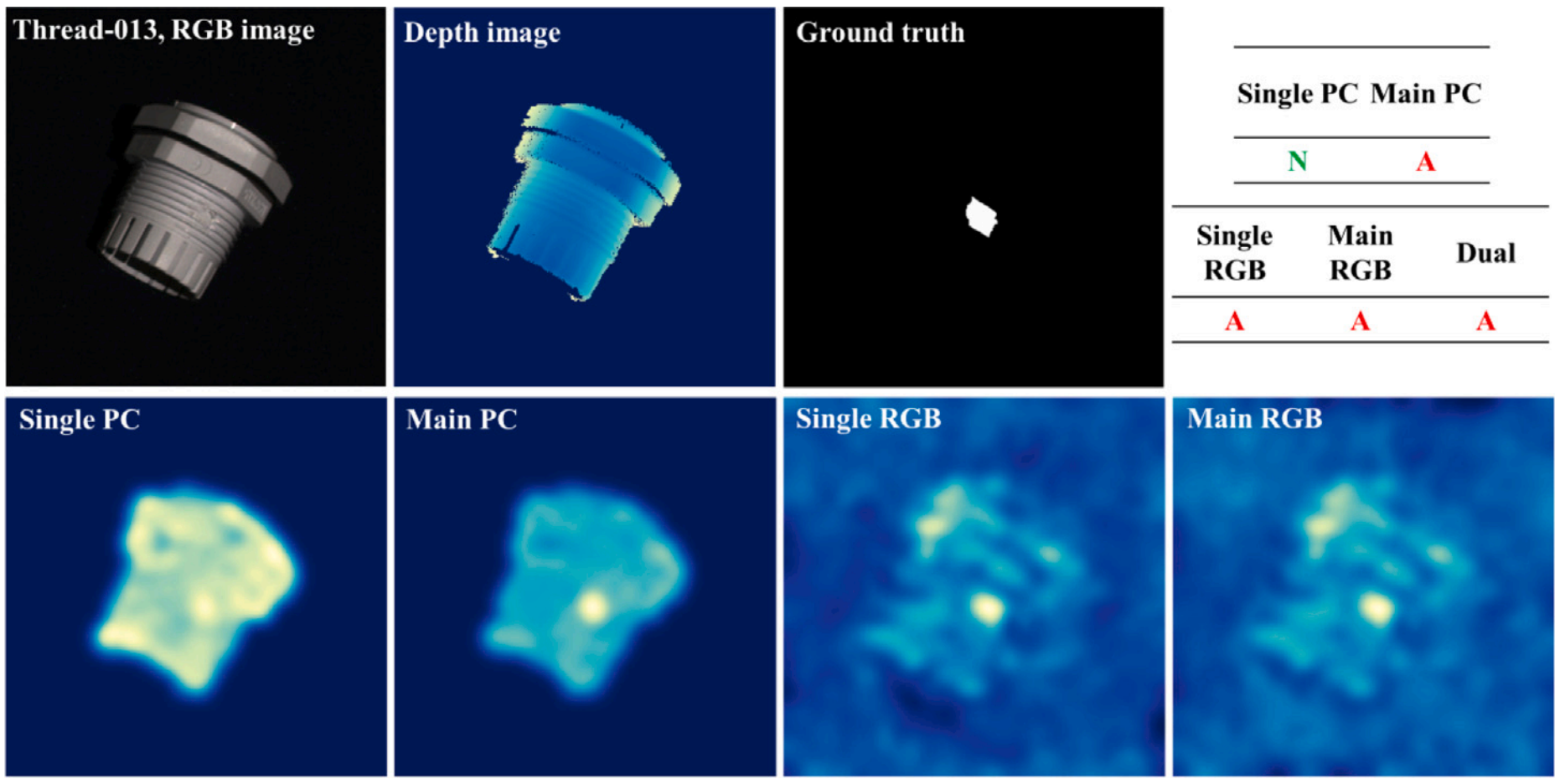

**Fig. A.1.** Visualization and prediction results of Cable Gland-Test-Thread-013. (A: Abnormal, N: Normal).

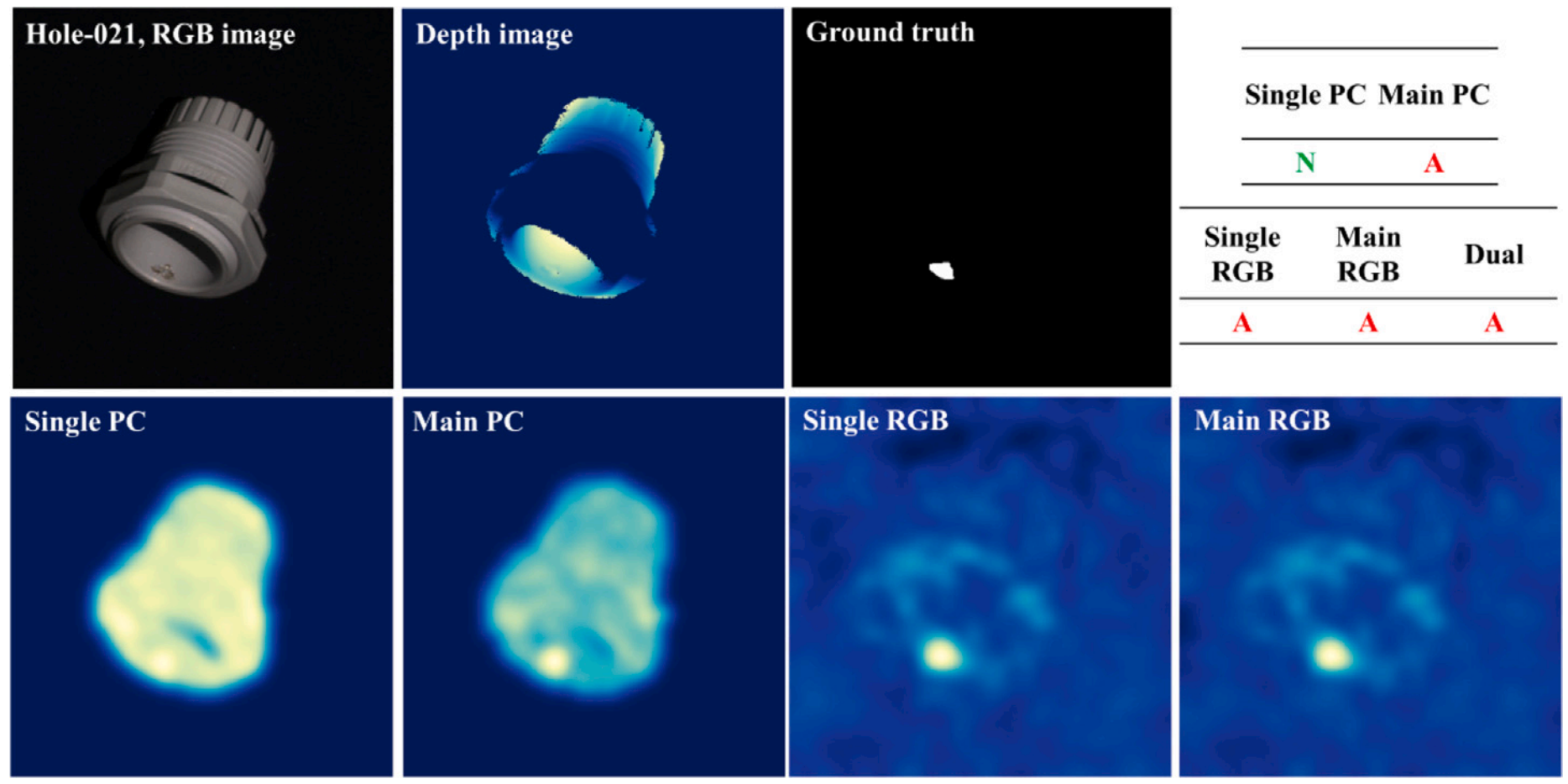

**Fig. A.2.** Visualization and prediction results of Cable Gland-Test-Hole-021. (A: Abnormal, N: Normal).

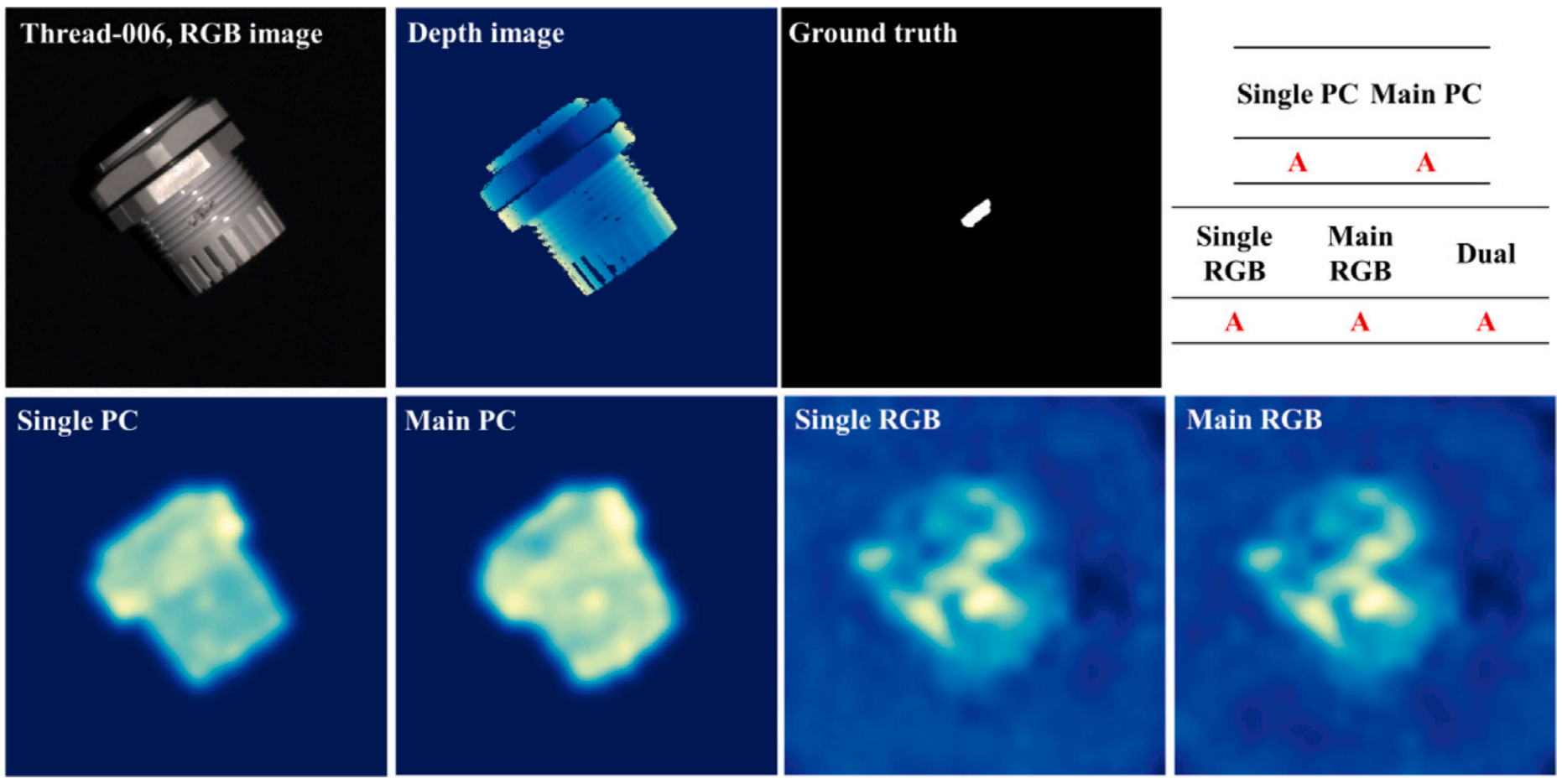

**Fig. A.3.** Visualization and prediction results of Cable Gland-Test-Thread-006. (A: Abnormal, N: Normal).

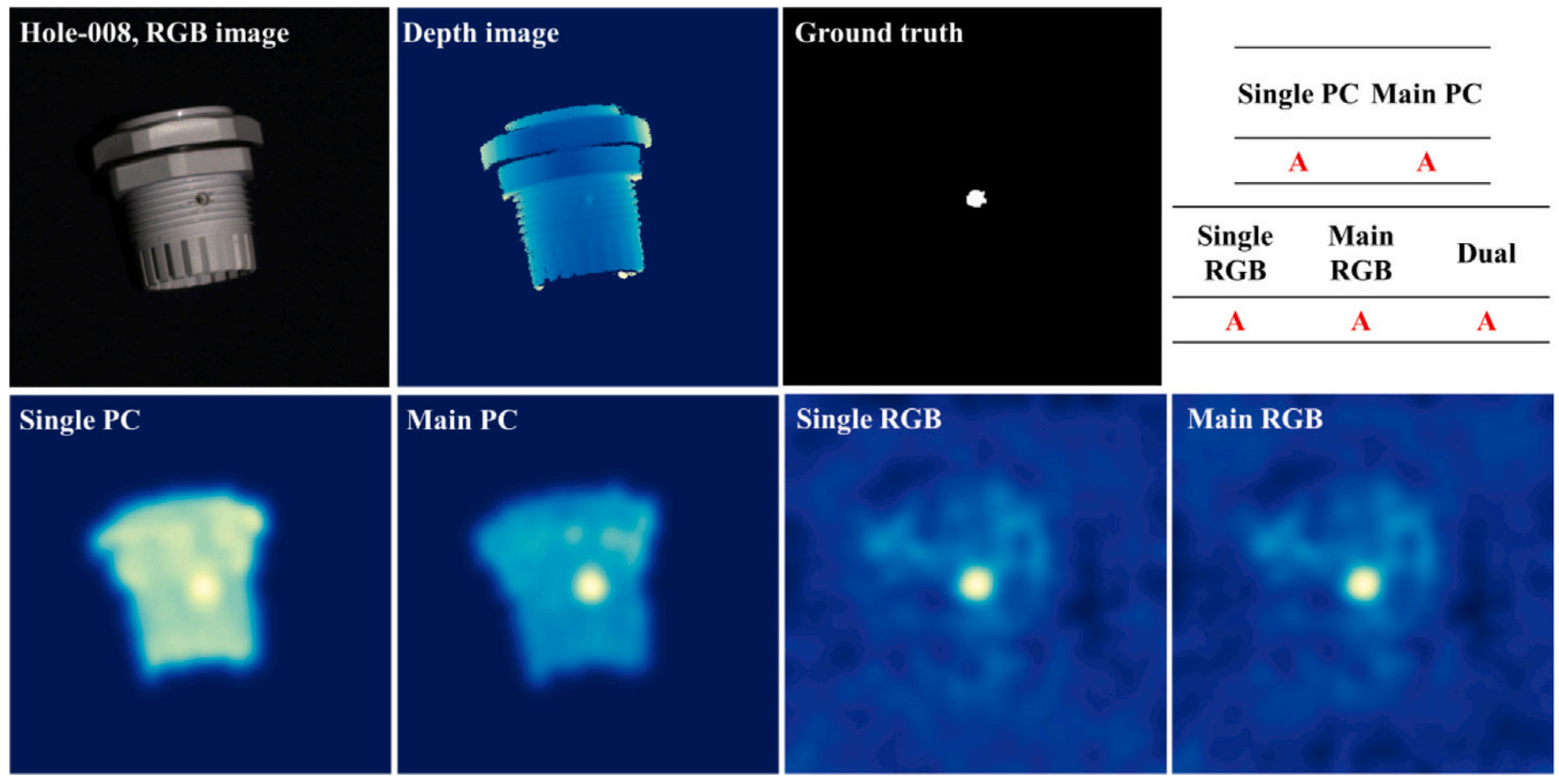

**Fig. A.4.** Visualization and prediction results of Cable Gland-Test-Hole-008. (A: Abnormal, N: Normal).

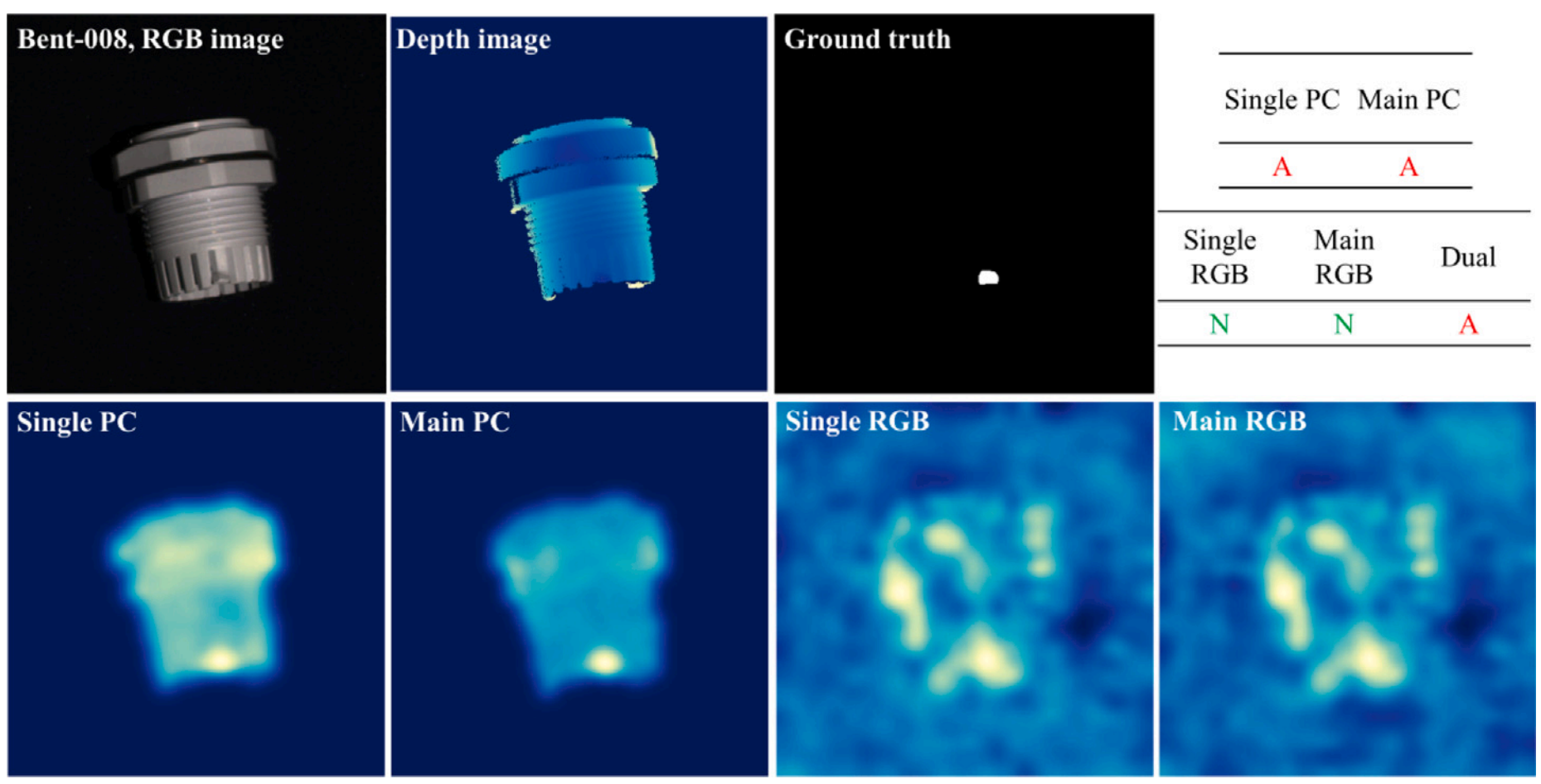

**Fig. A.5.** Visualization and prediction results of Cable Gland-Test-Bent-008. (A: Abnormal, N: Normal).

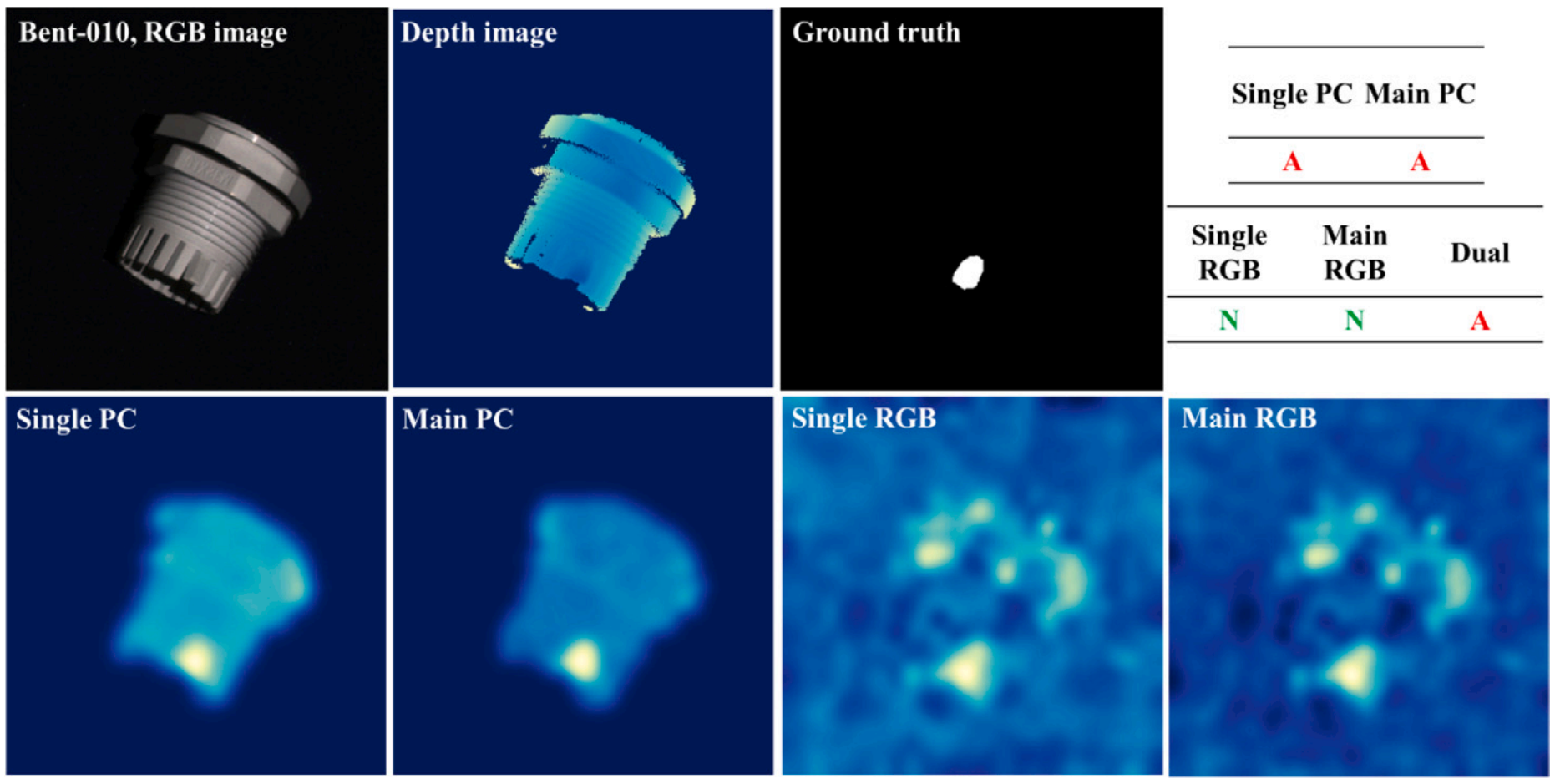

**Fig. A.6.** Visualization and prediction results of Cable Gland-Test-Bent-010. (A: Abnormal, N: Normal).

**Table A.1**
P-AUROC results on MVTec 3D-AD.

| | Method | Bagel | CableGland | Carrot | Cookie | Dowel | Foam | Peach | Potato | Rope | Tire | Mean |
|---|---|---|---|---|---|---|---|---|---|---|---|---|
| 3D | FPFH | 0.994 | 0.966 | 0.999 | 0.946 | 0.966 | 0.927 | 0.996 | 0.999 | 0.996 | 0.99 | 0.978 |
| | M3DM | 0.981 | 0.949 | 0.997 | 0.932 | 0.959 | 0.925 | 0.989 | 0.995 | 0.994 | 0.981 | 0.97 |
| | $\text{Ours}_{Dual\,Banks\,PCs}$ | 0.983 | 0.95 | 0.997 | 0.932 | 0.959 | 0.94 | 0.991 | 0.996 | 0.994 | 0.984 | 0.973 |
| | $\text{Ours}_{FtoF}$ | 0.992 | 0.975 | 0.999 | 0.96 | 0.975 | 0.924 | 0.998 | 0.998 | 0.998 | 0.996 | 0.981 |
| RGB | PatchCore | 0.983 | 0.984 | 0.98 | 0.974 | 0.972 | 0.849 | 0.976 | 0.983 | 0.987 | 0.977 | 0.967 |
| | M3DM | 0.992 | 0.99 | 0.994 | 0.977 | 0.983 | 0.955 | 0.994 | 0.99 | 0.995 | 0.994 | 0.987 |
| | $\text{Ours}_{Dual\,Banks\,RGB}$ | 0.992 | 0.993 | 0.994 | 0.977 | 0.983 | 0.956 | 0.993 | 0.99 | 0.995 | 0.994 | 0.987 |
| | $\text{Ours}_{FtoF}$ | 0.992 | 0.993 | 0.994 | 0.976 | 0.983 | 0.956 | 0.993 | 0.99 | 0.995 | 0.994 | 0.987 |
| RGB+3D | AST | – | – | – | – | – | – | – | – | – | – | 0.976 |
| | PatchCore | 0.996 | 0.992 | 0.997 | 0.994 | 0.981 | 0.974 | 0.996 | 0.998 | 0.994 | 0.995 | 0.992 |
| | M3DM | 0.995 | 0.993 | 0.997 | 0.985 | 0.985 | 0.984 | 0.996 | 0.994 | 0.997 | 0.996 | 0.992 |
| | $\text{Ours}_{dual}$ | 0.995 | 0.993 | 0.996 | 0.976 | 0.984 | 0.988 | 0.996 | 0.995 | 0.997 | 0.996 | 0.992 |

**Table A.2**
Comparison of teacher model and trained/untrained student model (F2F).

| | PCs bank only (Baseline) | RGB bank only (Teacher with different modal input) | F2F with PCs input only (Student untrained) | F2F with PCs input only (Student Trained) |
|---|---|---|---|---|
| I-AUROC | 0.860 | 0.851 | 0.868 | 0.938 |
| AUPRO | 0.911 | 0.942 | 0.912 | 0.934 |

**Table A.3**
I-AUROC and AUPRO results on MVTec 3D-AD for RGB+PCs based methods.

| | Method | Bagel | CableGland | Carrot | Cookie | Dowel | Foam | Peach | Potato | Rope | Tire | Mean |
|---|---|---|---|---|---|---|---|---|---|---|---|---|
| I-AUROC | 3D-ST | 0.950 | 0.483 | 0.986 | 0.921 | 0.905 | 0.632 | 0.945 | 0.988 | 0.976 | 0.542 | 0.833 |
| | PatchCore+FPFH | 0.918 | 0.748 | 0.967 | 0.883 | 0.932 | 0.582 | 0.896 | 0.912 | 0.921 | 0.886 | 0.865 |
| | AST | 0.983 | 0.873 | 0.976 | 0.971 | 0.932 | 0.885 | 0.974 | 0.981 | 1.000 | 0.797 | 0.937 |
| | M3DM | 0.994 | 0.909 | 0.972 | 0.976 | 0.960 | 0.942 | 0.973 | 0.899 | 0.972 | 0.850 | 0.945 |
| | Shape-guided | 0.986 | 0.894 | 0.983 | 0.991 | 0.976 | 0.857 | 0.990 | 0.965 | 0.960 | 0.869 | 0.947 |
| | Ours | 0.992 | 0.893 | 0.977 | 0.960 | 0.953 | 0.883 | 0.950 | 0.937 | 0.943 | 0.893 | 0.938 |
| AUPRO | 3D-ST | 0.950 | 0.483 | 0.986 | 0.921 | 0.905 | 0.632 | 0.945 | 0.988 | 0.976 | 0.542 | 0.833 |
| | PatchCore+FPFH | 0.976 | 0.969 | 0.979 | 0.973 | 0.933 | 0.888 | 0.975 | 0.981 | 0.950 | 0.971 | 0.959 |
| | M3DM | 0.970 | 0.971 | 0.979 | 0.950 | 0.941 | 0.932 | 0.977 | 0.971 | 0.971 | 0.975 | 0.964 |
| | Shape-guided | 0.981 | 0.973 | 0.982 | 0.971 | 0.962 | 0.978 | 0.981 | 0.983 | 0.974 | 0.975 | 0.976 |
| | Ours | 0.970 | 0.971 | 0.977 | 0.932 | 0.934 | 0.946 | 0.978 | 0.970 | 0.970 | 0.974 | 0.962 |

**Table A.4**
I-AUROC and AUPRO results based on MTFI pipeline (FtoF, PCs as main modality) with Different distance metrics.

| | Method | Bagel | CableGland | Carrot | Cookie | Dowel | Foam | Peach | Potato | Rope | Tire | Mean |
|---|---|---|---|---|---|---|---|---|---|---|---|---|
| I-AUROC | L2 | 0.992 | 0.893 | 0.977 | 0.960 | 0.953 | 0.883 | 0.950 | 0.937 | 0.943 | 0.893 | 0.938 |
| | L1 | 0.986 | 0.823 | 0.994 | 0.987 | 0.960 | 0.786 | 0.992 | 0.865 | 0.947 | 0.944 | 0.928 |
| | Cosine | 0.994 | 0.795 | 0.993 | 0.983 | 0.932 | 0.754 | 0.995 | 0.894 | 0.921 | 0.953 | 0.921 |
| P-AUROC | L2 | 0.992 | 0.975 | 0.999 | 0.960 | 0.975 | 0.924 | 0.998 | 0.998 | 0.998 | 0.996 | 0.981 |
| | L1 | 0.992 | 0.972 | 0.999 | 0.960 | 0.975 | 0.927 | 0.997 | 0.998 | 0.997 | 0.995 | 0.981 |
| | Cosine | 0.993 | 0.973 | 0.998 | 0.977 | 0.982 | 0.915 | 0.998 | 0.998 | 0.997 | 0.996 | 0.983 |
| AUPRO | L2 | 0.968 | 0.907 | 0.982 | 0.933 | 0.918 | 0.726 | 0.982 | 0.983 | 0.969 | 0.974 | 0.934 |
| | L1 | 0.968 | 0.897 | 0.982 | 0.931 | 0.917 | 0.727 | 0.981 | 0.982 | 0.968 | 0.971 | 0.932 |
| | Cosine | 0.972 | 0.900 | 0.981 | 0.956 | 0.940 | 0.694 | 0.982 | 0.983 | 0.972 | 0.973 | 0.935 |

### *A.4. Prediction results of ‘hole’ and ‘combined’ anomalies from Cookie class*

Prediction Results of ‘hole’ Fig. A.7 and ‘combined’ Fig. A.8 anomalies from Cookie class are also visualized here to support ideas in Section Analysis of Asymmetry Results of Cross-Modal Distillation.

### *A.5. Experiments compute resources*

We conducted experiments on a single NVIDIA RTX A6000 with Linux driver 535.54.03. The CPU is INTEL W5-2455X with 256 G DRR5 memory. For DualBanksPCs/RGB, it needs about 50 and 70 min and maximum 50 GB memory for RGB and PCs, respectively. For training the distillation network, it takes about 30 s to 1 min per epoch (TF32 is used). Since the step of transferring tensors to GPU memory in PyTorch is time-consuming, we loaded the data into GPU memory in advance to speed up the calculation. Running the MTFI pipeline with the trained distillation network takes approximately two hours. Unfortunately, training the distillation network is an intermediate step, and lower loss does not bring the best results to the MTFI pipeline. We need to run the MTFI pipeline multiple times with different checkpoints to find the optimized point. This also means that the general way of calculating per-pixel distance may need to be improved. For single memory bank and dual memory bank methods, you only need to run it once to get a result of test set defect inspection. For the two-stage MTFI pipeline, the cross-modal KD network needs to be trained first and then inference needs to be performed (i.e., it needs to be run twice). Choosing the appropriate number of epochs for training the cross-modal KD network requires repeatedly running the second stage using a grid search, and we searched in units of 5 epochs.

### *A.6. Inference speed*

We also measured inference speed of our MTFI pipeline (F2F, Inference with PCs) with similar methods. It can be seen that the inference time is basically in line with the expectations of the network structure. Due to the absence of feature fusion network and the omission of RGB feature extraction network, our method is slightly faster than M3DM. Although Crossmodal feature mapping also uses the memory bank method, it is significantly faster. This is partly due to network improvements, but more because some CPU calculations are transferred to the GPU (see Table A.5).

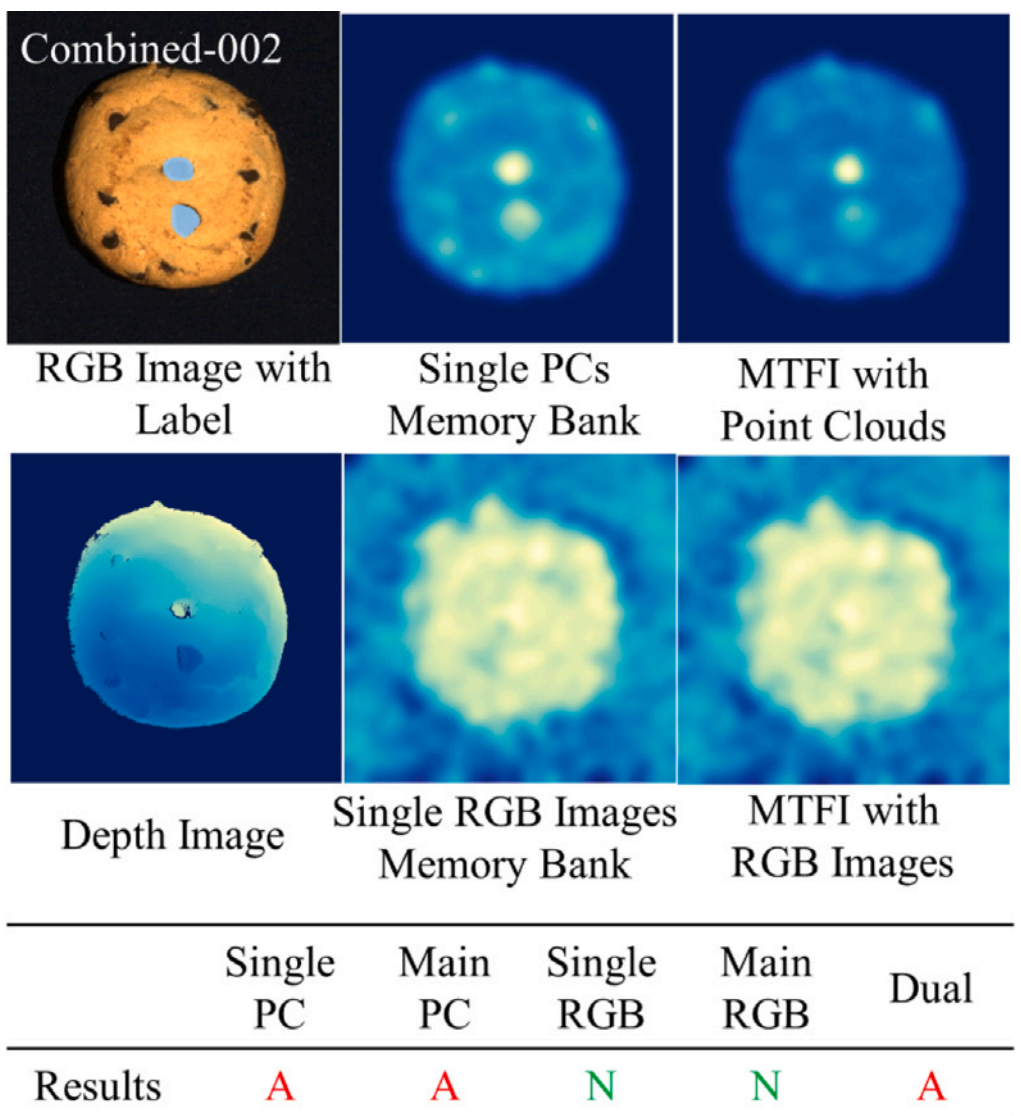

**Fig. A.7.** Visualization and prediction results of 'hole' anomalies from Cookie class. (A: Abnormal, N: Normal).

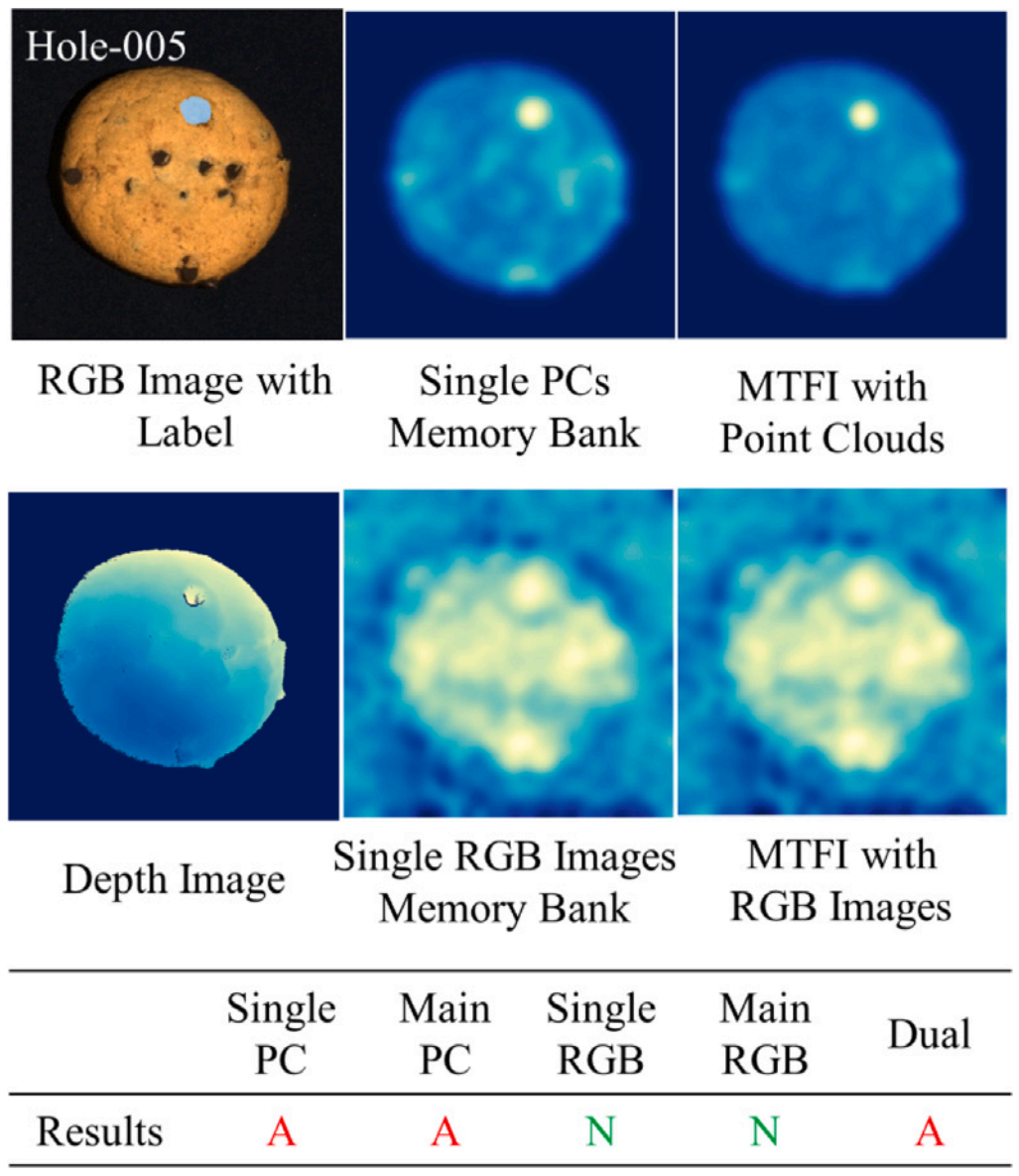

**Fig. A.8.** Visualization and prediction results of 'combined' anomalies from Cookie class. (A: Abnormal, N: Normal).

## Data availability

All used public datasets are cited within the article.

**Table A.5**
Comparison of inference times.

| Method | Inference time (s) |
|---|---|
| BTF [44] | 2.19 |
| Shape-guided [25] | 2.05 |
| M3DM [8] | 1.95 |
| Crossmodal feature mapping [6] | 0.14 |
| Ours (Dual RGB+PCs memory bank) | 1.90 |
| Ours (MTFI, F2F, Inference with PCs only) | 1.74 |